\documentclass[letterpaper,journal]{IEEEtran}

\usepackage{amsmath,amsfonts,amssymb}
\usepackage{array}
\usepackage{tabularx}
\usepackage{textcomp}
\usepackage{url}
\usepackage{graphicx}
\usepackage{bm}
\usepackage{cite}
\usepackage{booktabs}
\usepackage{array}
\usepackage{hyperref}

\hypersetup{
    colorlinks=true,
    linkcolor=blue,
    citecolor=blue,
    urlcolor=blue
}

\newcommand{\vect}[1]{\boldsymbol{#1}}

\newcommand{\FloatBarrier}{\relax}
\newcounter{algorithm}
\renewcommand{\thealgorithm}{\arabic{algorithm}}
\renewcommand{\arraystretch}{1.08}
\graphicspath{{figures/}}
\def\BibTeX{{\rm B\kern-.05em{\sc i\kern-.025em b}\kern-.08em
    T\kern-.1667em\lower.7ex\hbox{E}\kern-.125emX}}

\begin{document}

\title{SIGN-MAPPO: State–Intent-Aware Graph Multi-Agent Reinforcement Learning for Cooperative Driving under Imperfect V2X}

\newcommand{\paperpdfauthors}
{Zide Fan, Member, IEEE, Jiahui Yuan, Kaixin Zhang, Junhao Fan, Xiaohe Li, Yiru Wang}

\author{
Zide Fan,~\IEEEmembership{Member,~IEEE}, Jiahui Yuan, Kaixin Zhang, Junhao Fan, Xiaohe Li, Yiru Wang%
\thanks{Zide Fan, Xiaohe Li, Yiru Wang are with the Aerospace Information Research Institute, Chinese Academy of Sciences, Beijing 100190, China, and also with the Key Laboratory of Target Cognition and Application Technology, Aerospace Information Research Institute, Chinese Academy of Sciences, Beijing, China (fanzd@aircas.ac.cn; lixiaohe@aircas.ac.cn; yiruwang@aircas.ac.cn).}%
\thanks{Jiahui Yuan, Kaixin Zhang, Junhao Fan are with the Aerospace
Information Research Institute, Chinese Academy of Sciences,
and the Key Laboratory of Target Cognition and Application
Technology (TCAT); they are also with the University of
Chinese Academy of Sciences and the School of Electronic,
Electrical and Communication Engineering, University of
Chinese Academy of Sciences (yuanjiahui26@mails.ucas.ac.cn; zhangkaixin25@mails.ucas.ac.cn; fanjunhao25@mails.ucas.ac.cn).}%

}

\maketitle

\begin{abstract}
Cooperative driving at mixed-traffic unsignalized intersections is challenged by imperfect vehicle-to-everything (V2X) communication, human-driven vehicle (HV) intentions that cannot be directly communicated, and the potential incompatibility of independently generated crossing decisions. This paper proposes State--Intent-Aware Graph Multi-Agent Reinforcement Learning for Cooperative Driving under Imperfect V2X (SIGN-MAPPO) to address these challenges through three coordinated mechanisms. Each decision-eligible connected and automated vehicle (CAV) first reconstructs a receiver-specific traffic graph from its ego state and the peer-state records available to it, from which a shared graph-attention policy generates a raw \textit{Stop/Go} intention augmented with predicted HV-intent features. CAV intentions are then exchanged through timestamped vehicle-to-vehicle (V2V) messages, and a local conflict-aware arbitration layer converts each CAV's intention into a safe executable action despite potentially inconsistent peer information. In the principal Simulation of Urban MObility (SUMO) evaluation, SIGN-MAPPO achieves a 97.53\% completion rate, a 0.10\% collision rate, and a mean intersection traversal time of 35.28~s. Ablation studies further confirm the complementary contributions of HV-intent prediction and CAV-intent exchange to this performance. The trained policy also remains robust under the evaluated communication-delay and packet-loss ranges, and cross-scenario tests show complete traffic clearance across the tested settings, though safety margins narrow under compound unpredictable HV behavior.
\end{abstract}

\begin{IEEEkeywords}
Connected and automated vehicles, graph attention network, human-driven vehicle intent, multi-agent reinforcement learning, unsignalized intersection, vehicle-to-everything communication.
\end{IEEEkeywords}

\section{Introduction}
\label{sec:introduction}

\IEEEPARstart{I}{ntersections} are essential to urban road
networks, yet concentrated traffic interactions make them
bottlenecks for safety and mobility. As connected and
automated vehicles (CAVs) enter transportation systems,
they are expected to coexist with human-driven vehicles
(HVs) for an extended period \cite{mahmassani2016mixed}.
CAVs can communicate with one another, whereas conventional
HVs cannot directly exchange traffic information and exhibit
heterogeneous driving behavior \cite{wang2023mixedcarfollowing}.
These differences complicate the anticipation and
coordination of crossings at unsignalized intersections.

Cooperative crossing depends in part on vehicle-state
information exchanged through vehicle-to-everything (V2X) communication. Many
intersection methods use a shared or synchronized traffic
state \cite{guan2020centralized,cai2022multilane}, but
transmission delay and packet loss can deliver different
updates to different CAVs or prevent some updates from
arriving \cite{yu2022communicationdelay}. CAVs may therefore
hold different information about the same surrounding vehicle,
making their crossing decisions harder to coordinate.

Vehicle states alone do not describe how traffic participants
intend to move. Existing mixed-traffic studies generally do
not distinguish CAV intentions from HV intentions in their
decision representations
\cite{zhuang2023cooperative,wang2025mixed}. CAV intentions
can be communicated directly, whereas HV intentions must be
inferred from motion histories and remain uncertain. These
information sources call for separate representations.
Moreover, even accurate state and intention information does
not ensure compatible actions when CAVs make decentralized
decisions about conflicting movements. The combined problem
of heterogeneous intentions and decentralized action
compatibility remains insufficiently addressed.

Existing approaches address these challenges only partially. Rule-based and reservation-based methods~\cite{dresner2008multiagent,levin2017conflict} provide interpretable coordination but generally assume synchronized or complete vehicle-state access, offering limited tolerance to inconsistent receiver-side information. Learning-based single-agent and multi-agent approaches~\cite{isele2018occluded,alsharman2023hierarchical,huang2023expert,crosato2023svo,zhang2020guided,li2023transformer,fu2022multimodal,zhang2024interactive,li2024gaming,tang2022sac} improve adaptability to interactive traffic but typically encode HV behavior implicitly within a shared state representation, without separating inferred HV intent from directly communicable CAV intent. Recent MARL-based cooperative-driving methods~\cite{zhuang2023cooperative,wang2025mixed} further improve coordination among CAVs but do not explicitly reconcile independently generated crossing decisions when V2X communication is imperfect. Consequently, no existing framework jointly addresses receiver-dependent state inconsistency, the distinct communicability of CAV and HV intentions, and the compatibility of decentralized decisions---the three factors that jointly determine whether cooperative crossing decisions remain safe and efficient once assumptions of synchronized communicable information are relaxed.

To address these gaps, this paper proposes
State--Intent-Aware Graph Multi-Agent Reinforcement Learning
for Cooperative Driving under Imperfect V2X (SIGN-MAPPO).
Each CAV builds a receiver-specific traffic graph from the
vehicle states available to it, with HV nodes augmented by
intent features inferred from receiver-visible motion
histories. A shared graph-attention policy generates raw CAV
crossing intentions, which are exchanged through vehicle-to-vehicle (V2V)
communication. Each CAV then applies conflict-aware local
arbitration to determine its executable action. Multi-agent proximal policy optimization (MAPPO)
\cite{yu2022mappo} trains the policy under centralized
training and decentralized execution (CTDE)
\cite{lowe2017maddpg}, while decisions remain local during
execution.

The main contributions of this study are summarized as
follows:

\begin{enumerate}
\renewcommand{\labelenumi}{\arabic{enumi})}
    \item A receiver-specific graph decision model is developed
    for cooperative driving under imperfect V2X communication.
    Each CAV constructs its own traffic graph from the
    vehicle-state information available to that receiver,
    yielding different graph representations when the received
    information differs across CAVs.

    \item A decoupled HV- and CAV-intent modeling scheme is
    introduced. HV intentions are predicted from their observed
    motion histories and added to the corresponding HV node
    features, while CAV intentions are explicitly exchanged
    through a separate V2V communication path.

    \item A conflict-aware local arbitration mechanism is
    designed for decentralized execution. Each CAV locally
    coordinates its own intention with the received intentions
    of other CAVs and converts the resulting decision into an
    executable action, improving the compatibility of actions
    selected by different CAVs.

    \item Four groups of experiments are conducted, comprising baseline comparisons, module ablations, communication-robustness tests, and cross-scenario evaluations. The results show that SIGN-MAPPO achieves a 97.53\% completion rate and a 0.10\% collision rate while maintaining robust performance under communication impairments and across diverse traffic scenarios.
\end{enumerate}

The remainder of this paper is organized as follows.
Section~\ref{sec:related_work} reviews unsignalized-intersection
coordination, learning-based mixed-traffic decision making,
graph and intent representations, and multi-agent learning
under communication uncertainty. Section~\ref{sec:problem_statement}
defines the mixed-traffic crossing problem and its optimization
objective. Section~\ref{sec:method} presents the proposed
SIGN-MAPPO decision model and its main components.
Section~\ref{sec:experiments} reports the comparative,
ablation, communication-robustness, and cross-scenario
experiments. Section~\ref{sec:conclusion} concludes the paper.

\section{Related Work}
\label{sec:related_work}

\subsection{Unsignalized-Intersection Coordination}

Early autonomous intersection management relies on explicit right-of-way rules, reservations, or conflict-point constraints. Dresner and Stone \cite{dresner2008multiagent} formulated reservation-based intersection management, and Levin and Rey \cite{levin2017conflict} optimized reservation timing through conflict-point separation. These methods provide interpretable safety constraints but generally require accurate vehicle states and coordinated access to the intersection. Formation-based and centralized cooperative controllers extend this line of work to connected vehicles \cite{cai2022multilane,guan2020centralized}. Mixed-traffic studies have examined cyber interactions among vehicles \cite{guo2023mixed}; basic safety messages have been used to study driving decisions \cite{kamrani2020mdp}, while individual car-following styles inform cruise control design \cite{chu2021cruise}. Recent intersection policies address random tasks and lane-graph structure \cite{xiao2024random,liu2024lanegraph}, but their performance still depends on the communication and controllability assumptions used to construct a common scene.

\subsection{Learning-Based Mixed-Traffic Decision Making}

DRL has been used to learn gap acceptance, crossing, and cooperative release policies from interaction data, within a broader progression toward environment-aware driving decisions. Isele \emph{et al.} \cite{isele2018occluded} studied deep Q-learning at occluded intersections, while Al-Sharman \emph{et al.} \cite{alsharman2023hierarchical} combined high-level reinforcement learning with feasible motion control. Demonstration-aided learning improves sample efficiency and constrains unsafe exploration \cite{huang2023expert}, whereas social-value and uncertainty-aware formulations target interaction quality and branching efficiency \cite{crosato2023svo,zhang2020guided}. Transformer and multimodal representations have also been introduced for lane changing and urban driving decisions \cite{li2023transformer,fu2022multimodal}.

For multi-vehicle cooperation, Zhuang \emph{et al.} \cite{zhuang2023cooperative} introduced attention-based MAPPO with weighted local rewards. Wang \emph{et al.} \cite{wang2025mixed} developed a decentralized mixed-traffic controller and conflict-resolution mechanism for complex intersections. Related work has examined online interactive prediction, local-game reasoning, and soft actor--critic control \cite{zhang2024interactive,li2024gaming,tang2022sac}. These studies establish the value of learned coordination, but their traffic encodings do not jointly distinguish receiver-dependent state freshness and the different intent sources of CAVs and HVs.

\subsection{Graph and Intent Representations for Driving Policies}

Graph neural networks encode traffic participants as nodes and their interactions as edges, reducing the dependence on a fixed vector ordering. Graph attention networks (GATs) further learn neighbor importance through masked self-attention \cite{velickovic2018gat}. In autonomous driving, Ma \emph{et al.} \cite{ma2021latent} combined latent driver-state inference with spatial--temporal graph relations, Mo \emph{et al.} \cite{mo2022heterogeneous} modeled heterogeneous edges for trajectory prediction, and Ye \emph{et al.} \cite{ye2022gsan} learned spatial--temporal interaction representations using graph self-attention. Peng \emph{et al.} \cite{peng2022drlgat} coupled GAT-based planning with a safety architecture, and Zheng \emph{et al.} \cite{zheng2026gatdrl} used graph attention to capture multi-hop interaction dependencies at unsignalized intersections. Gated recurrent units (GRUs) provide a recurrent mechanism for encoding sequential observations \cite{cho2014gru}. SIGN-MAPPO builds on these components but assigns them specific information roles: graph nodes are constructed for each receiver, inferred HV intent enters the policy representation, and communicated CAV intent supports execution-side coordination.

\subsection{Multi-Agent Training Under Communication Uncertainty}

Centralized-training/decentralized-execution methods reduce multi-agent nonstationarity during optimization while preserving distributed policies at deployment \cite{lowe2017maddpg}. Counterfactual credit assignment and value decomposition provide alternative cooperative-learning mechanisms \cite{foerster2018coma,rashid2018qmix}. Proximal policy optimization (PPO) supplies a stable clipped policy objective \cite{schulman2017ppo}, and MAPPO has shown strong performance across cooperative multi-agent benchmarks \cite{yu2022mappo}. Vehicular deployment adds a communication layer in which a received record may be valid but stale. Age-of-information research \cite{kaul2012realtime} highlights the importance of status-update freshness, while delay and packet loss have been studied in signal-free intersection control \cite{yu2022communicationdelay}. These findings motivate explicit timestamp tracking rather than treating every delivered packet as current. SIGN-MAPPO incorporates this principle through directed latest-valid caches and evaluates state-only as well as joint state--intent delay and loss.

\section{Problem Statement}
\label{sec:problem_statement}

\begin{figure}[!t]
\centering
\includegraphics[width=\columnwidth]
{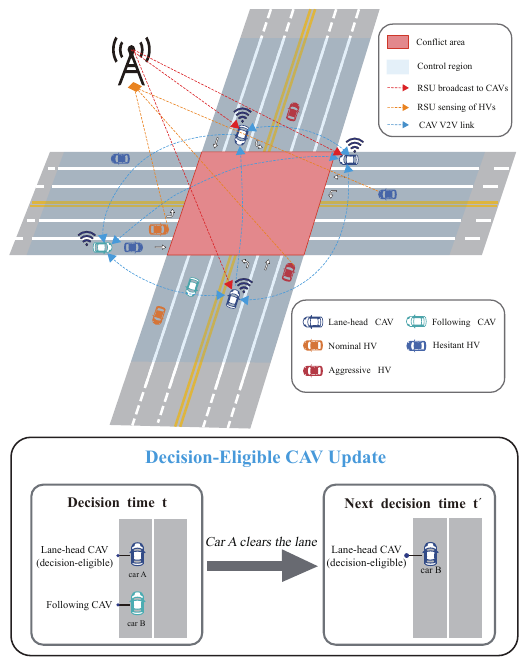}
\caption{Mixed-traffic unsignalized-intersection scenario,
V2V and RSU-assisted information flows, and the changing
set of decision-eligible CAVs.}
\label{fig:scenario_communication}
\end{figure}

We consider a four-leg unsignalized intersection with one
left-turn lane and two through lanes on each approach. The
area enclosed by the stop lines is the conflict area, and the
upstream segment of each approach is the control region.
Traffic comprises CAVs and HVs; the latter exhibit hesitant,
nominal, or aggressive driving styles with stochastic
within-style variation. CAVs exchange information via V2V
communication, while an RSU observes HV states and broadcasts
them to CAVs. The V2V range exceeds twice the control-region
length, allowing approaching CAVs to communicate before
entering the conflict area. Both communication paths are
subject to delay and packet loss. Fig.~\ref{fig:scenario_communication}
illustrates the scenario and information flows.

On each lane, the vehicle nearest the stop line is the lane-head
vehicle. A lane-head CAV inside the control region is
decision-eligible and receives a high-level crossing decision
from the proposed method. Following CAVs remain under
car-following control until becoming lane-head.

\subsection{Dec-POMDP Formulation}
\label{subsec:problem_formulation}

Traffic interaction at an unsignalized intersection is modeled
as a sequential decision-making problem. A decision made for
a lane-head CAV affects its subsequent motion and therefore the
traffic state at the next decision step. Multiple lane-head CAVs may
require decisions at the same time, and each CAV makes its
decision using receiver-specific and potentially incomplete
traffic information. The intersection-crossing problem is
therefore formulated as a decentralized partially observable
Markov decision process (Dec-POMDP).

The principal symbols used throughout this paper are summarized
in Table~\ref{tab:notation}. The Dec-POMDP is defined as
\begin{equation}
\mathcal M
=
\left\langle
\mathcal I,\mathcal S,
\{\mathcal O_i\},
\{\mathcal A_i\},
\mathcal P,\mathcal R,\gamma
\right\rangle,
\label{eq:dec_pomdp}
\end{equation}
where $\mathcal I$ denotes the set of CAVs considered during an
episode, and $\mathcal S$ is the complete traffic-state space.
The time-varying subset $\mathcal I_t\subseteq\mathcal I$,  contains the lane-head CAVs that
are eligible to make a new decision at step $t$. For each CAV
$i\in\mathcal I$,
$\mathcal O_i$ and $\mathcal A_i$ denote its observation and
action spaces, respectively. The transition function
$\mathcal P(s_{t+1}\mid s_t,\vect a^{\mathrm{exe}}_t)$ gives the
conditional distribution of the next traffic state under the
current traffic state and joint executable actions. The reward
function is denoted by $\mathcal R$, and $\gamma\in[0,1)$ is the
discount factor. The state
and observation spaces, action space, and reward-based control
objective are specified in the following subsections.

\begin{table}[!t]
\caption{Principal Notation Used in the Problem Formulation}
\label{tab:notation}
\centering
\scriptsize
\setlength{\tabcolsep}{3pt}
\renewcommand{\arraystretch}{1.08}
\begin{tabular}{@{}p{0.25\columnwidth}p{0.69\columnwidth}@{}}
\toprule
\textbf{Symbol} & \textbf{Definition}\\
\midrule
$t$, $T$, and $\Delta t$ & Decision-step index, finite horizon, and decision interval\\
$i$ and $j$ & Index of a decision-eligible CAV $i\in\mathcal I_t$ and index of any vehicle $j$ represented in the traffic state, respectively\\
$R_{\mathrm{ctrl}}$ & Longitudinal length of the control region\\
$\mathcal I$ and $\mathcal I_t$ & CAVs considered in an episode and decision-eligible lane-head CAVs at step $t$\\
$\mathcal S$ and $s_t$ & Complete traffic-state space and its realized state at step $t$\\
$\mathcal P$ and $\mathcal R$ & State-transition function and reward function\\
$\gamma$ & Discount factor\\
$\vect s_{j,t}$ & State record of vehicle $j$ at step $t$\\
$\vect p_{j,t}$; $x_{j,t},y_{j,t}$ & Planar position and its Cartesian coordinates\\
$v_{j,t}$ and $a^{\mathrm{acc}}_{j,t}$ & Longitudinal speed and acceleration\\
$d_{j,t}$ and $w_{j,t}$ & Stop-line distance and accumulated waiting time\\
$\ell_{j,t}$ and $\mu_j$ & Occupied lane and predefined movement direction\\
$c_j$ and $\eta^{\mathrm{head}}_{j,t}$ & Vehicle-type and lane-head indicators\\
$\mathcal O_i$ and $o_{i,t}$ & Observation space and observation of CAV $i$\\
$\widetilde{\vect s}_{j\mid i,t}$ & Latest valid state record of vehicle $j$ available to CAV $i$\\
$\mathcal A_i$ & High-level action space of CAV $i$\\
$a^{\mathrm{raw}}_{i,t}$ and $a^{\mathrm{exe}}_{i,t}$ & Raw crossing intention and executable action of CAV $i$\\
$\vect a^{\mathrm{exe}}_t$ & Joint executable actions at step $t$\\
$r_{i,t}$ and $r^{\mathrm{traffic}}_t$ & Total reward of CAV $i$ and shared traffic reward\\
$\rho^{\mathrm{kin}}_{i,t}$ and $o^{\mathrm{conf}}_{i,t}$ & Kinematic HV-risk score and maximum route-conflict time overlap\\
$\boldsymbol\pi$ & Joint policy of the CAVs\\
\bottomrule
\end{tabular}
\end{table}

\subsection{State and Observation Space}
\label{subsec:state_observation}

The state space $\mathcal S$ is the set of all possible complete
traffic states in the modeled intersection. At decision step $t$,
$s_t\in\mathcal S$ denotes the current complete traffic state,
which consists of the state records of the vehicles present in
the intersection and its control regions.

The state record of vehicle $j$ at decision step $t$ is defined as
\begin{equation}
\vect s_{j,t}
=
\left(
\vect p_{j,t},
v_{j,t},
a^{\mathrm{acc}}_{j,t},
d_{j,t},
w_{j,t},
\ell_{j,t},
\mu_j,
c_j
\right),
\label{eq:vehicle_state}
\end{equation}
where $\vect p_{j,t}=[x_{j,t},y_{j,t}]^{\mathsf T}$ denotes the
planar position of vehicle $j$, $v_{j,t}$ is its longitudinal
speed, and $a^{\mathrm{acc}}_{j,t}$ is its longitudinal
acceleration. The variables $d_{j,t}$ and $w_{j,t}$ denote its
distance to the stop line and accumulated waiting time,
respectively. The lane occupied by the vehicle is denoted by
$\ell_{j,t}$, $\mu_j\in\{\mathrm{through},\mathrm{left}\}$
denotes its predefined movement direction, and $c_j$ indicates
whether it is a CAV or an HV.

Because state transmissions may be delayed or lost, CAV $i$
may not receive the current state $\vect s_{j,t}$ of another
vehicle $j$ at decision step $t$. We therefore use
$\widetilde{\vect s}_{j\mid i,t}$ to denote the most recently
received valid state record of vehicle $j$ available to CAV $i$.
The observation of CAV $i$ is then defined as
\begin{equation}
o_{i,t}
=
\left(
\vect s_{i,t},
\left\{
\widetilde{\vect s}_{j\mid i,t}
\right\}_{j\ne i}
\right)
\in\mathcal O_i,
\label{eq:receiver_observation}
\end{equation}
where $\mathcal O_i$ is the observation space of CAV $i$ and
$\vect s_{i,t}$ is its current state. Here, $j$ ranges only over
the other vehicles whose
state records are available to CAV $i$ at the current decision
step. If a newly generated state record is delayed or lost, the
previously received valid record is retained.

\subsection{Action Space}
\label{subsec:action_space}

At decision step $t$, the set of CAVs eligible to make an
intersection-crossing decision is defined as
\begin{equation}
\mathcal I_t
=
\left\{
i\in\mathcal I
\;\middle|\;
\eta^{\mathrm{head}}_{i,t}=1,\ 
0\le d_{i,t}\le R_{\mathrm{ctrl}}
\right\},
\label{eq:eligible_cav_set}
\end{equation}
where $\eta^{\mathrm{head}}_{i,t}=1$ indicates that CAV $i$ is
the lane-head vehicle, and $R_{\mathrm{ctrl}}$ is the
longitudinal length of the control region.

For each $i\in\mathcal I_t$, the action space is
\begin{equation}
\mathcal A_i
=
\{\mathrm{Stop},\mathrm{Go}\},
\qquad
a^{\mathrm{raw}}_{i,t},a^{\mathrm{exe}}_{i,t}\in\mathcal A_i.
\label{eq:action_space}
\end{equation}
The \emph{Stop} action requires the CAV to remain upstream of
the stop line, whereas the \emph{Go} action permits it to
proceed toward and enter the conflict area. These actions
represent high-level intersection-crossing decisions rather
than continuous vehicle-control inputs. Each eligible CAV first
generates a raw \emph{Stop}/\emph{Go} request
$a^{\mathrm{raw}}_{i,t}$, which is subsequently converted into
an executable action $a^{\mathrm{exe}}_{i,t}$ by the coordination
mechanism described in Section~IV. The policy produces the
joint raw requests, whereas the traffic-state transition is
determined by the corresponding joint executable actions.

\subsection{Reward Function and Control Objective}
\label{subsec:reward_objective}

Cooperative crossing requires the policy to improve vehicle
progress and intersection discharge without encouraging unsafe
entry or treating necessary yielding as inefficient. The reward
function therefore combines seven vehicle-specific components---
passage, progress, delay, yielding, arbitration, risk, and
collision---with an intersection-level traffic component. The
vehicle-specific components evaluate the crossing process of each
decision-eligible CAV, while the traffic component evaluates
intersection discharge and queue accumulation. Their activation
conditions and effects are summarized in
Table~\ref{tab:reward_components}.

\begin{table}[!t]
\caption{Environmental Reward Components}
\label{tab:reward_components}
\centering
\scriptsize
\setlength{\tabcolsep}{2.5pt}
\renewcommand{\arraystretch}{1.12}
\begin{tabular}{@{}>{\raggedright\arraybackslash}p{0.20\columnwidth}>{\raggedright\arraybackslash}p{0.73\columnwidth}@{}}
\toprule
\textbf{Component} & \textbf{Activation and effect}\\
\midrule
Passage & Positive when a CAV exits the conflict area; decreases as its accumulated waiting time increases\\
Progress & Positive when a CAV moves toward the stop line and negative when it moves away; the magnitude is bounded\\
Delay & Nonpositive before conflict-area entry; increases in magnitude with elapsed time, near-stop blocking, excessive waiting, new stops, and low-speed motion\\
Yielding & Positive when a CAV yields to a longer conflicting queue; includes an adjustment for the kinematic risk of conflicting HVs\\
Arbitration & Positive when a raw \emph{Go} intention is retained and negative when it is changed to \emph{Stop}\\
Risk & Negative when a raw \emph{Go} intention violates the waiting condition or temporally overlaps a conflicting route\\
Collision & Negative when a collision involving the CAV is newly detected; zero otherwise\\
Traffic & Positive for newly cleared vehicles and negative for queued vehicles at the intersection level\\
\bottomrule
\end{tabular}
\end{table}

The total reward assigned to CAV $i$ is
\begin{equation}
\begin{aligned}
r_{i,t}
={}&r^{\mathrm{veh}}_{i,t}
+r^{\mathrm{traffic}}_t,\\
r^{\mathrm{veh}}_{i,t}
={}&r^{\mathrm{pass}}_{i,t}
+r^{\mathrm{prog}}_{i,t}
+r^{\mathrm{delay}}_{i,t}
+r^{\mathrm{yield}}_{i,t}
&+r^{\mathrm{arb}}_{i,t}
+r^{\mathrm{risk}}_{i,t}
+r^{\mathrm{col}}_{i,t}.
\end{aligned}
\label{eq:total_reward}
\end{equation}
The passage, progress, delay, yielding, arbitration, risk, collision, and traffic components
follow the event-based design given in
Table~\ref{tab:reward_components}.

Passage and collision outcomes observed after a CAV enters the
conflict area are attributed to its preceding crossing decision.
Because the traffic reward is shared, it is counted once rather
than once for every active CAV. Let
$\bar r^{\mathrm{veh}}_t$ denote the mean vehicle-specific reward
over $\mathcal I_t$, with $\bar r^{\mathrm{veh}}_t=0$ when
$\mathcal I_t=\varnothing$. The control objective is
\begin{equation}
\boldsymbol\pi^{*}
=
\arg\max_{\boldsymbol\pi}
\mathbb E_{\boldsymbol\pi}
\left[
\sum_{t=0}^{T-1}\gamma^t
\left(
\bar r^{\mathrm{veh}}_t
+r^{\mathrm{traffic}}_t
\right)
\right],
\label{eq:decision_objective}
\end{equation}
where $\boldsymbol\pi$ maps the CAV observations to raw crossing
intentions. The expectation is taken over the traffic evolution
induced by the corresponding executable actions.

\section{Proposed SIGN-MAPPO Decision Model}
\label{sec:method}

\subsection{Model Framework}

\begin{figure*}[!t]
\centering
\includegraphics[width=\textwidth]{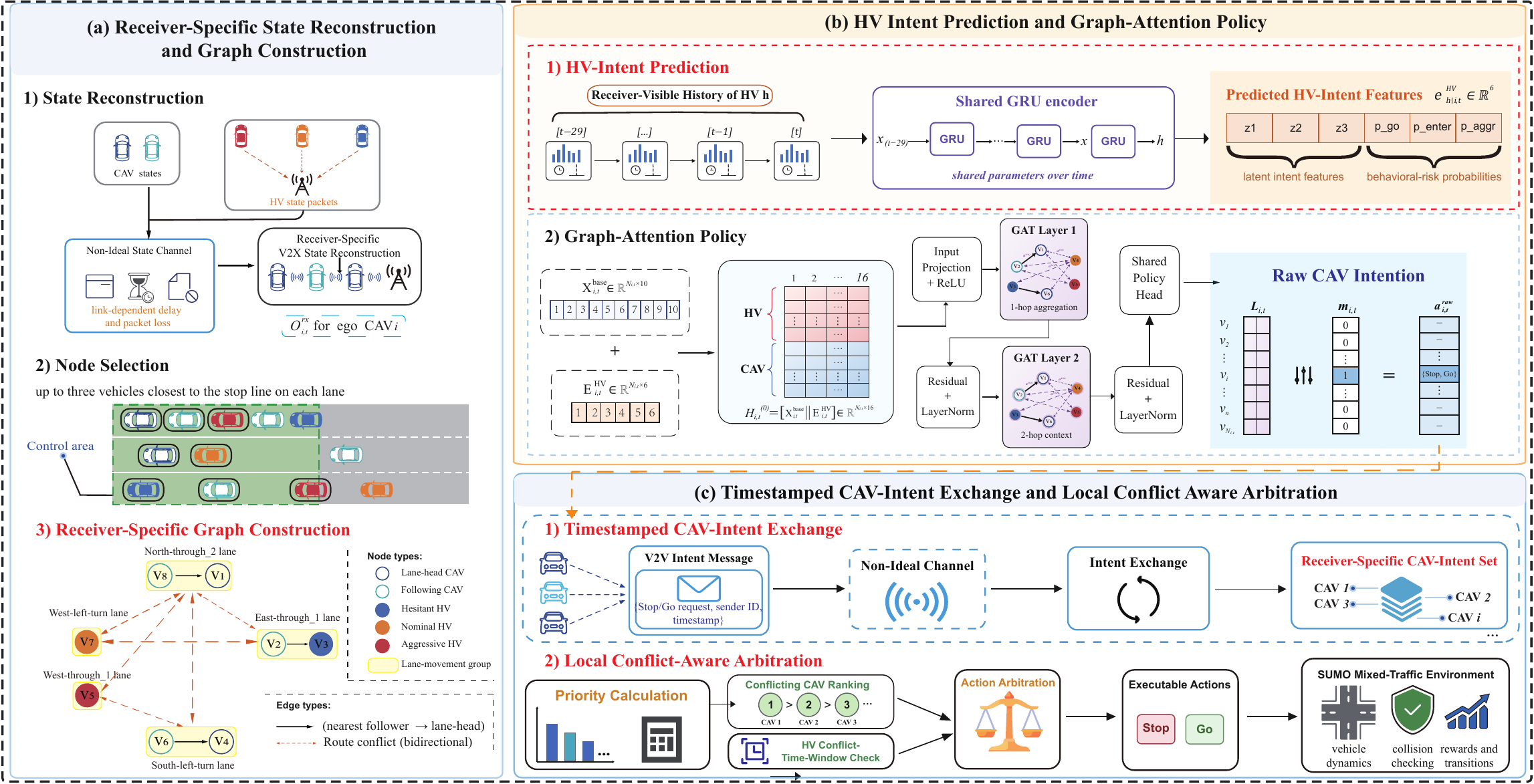}
\caption{Online decision process of SIGN-MAPPO:
(a) receiver-specific state reconstruction and graph
construction, (b) HV-intent prediction and graph-attention
policy, and (c) timestamped CAV-intent exchange and local
conflict-aware arbitration.}
\label{fig:sign_mappo_framework}
\end{figure*}

Decision making at mixed-traffic unsignalized intersections is
challenged by three coupled uncertainties. Imperfect V2X
communication leads to delayed, missing, and receiver-dependent
vehicle-state information. Meanwhile, HV intentions are not
directly observable and must be inferred from their motion
histories. In addition, independently generated CAV decisions
may become mutually incompatible when route-conflicting
vehicles simultaneously request access, making reliable
decentralized coordination difficult. To address these coupled challenges, this study proposes State–Intent-Aware Graph Multi-Agent Reinforcement Learning for Cooperative Driving under Imperfect V2X (SIGN-MAPPO), a cooperative crossing-decision framework for CAVs. As shown in Fig.~\ref{fig:sign_mappo_framework}, the framework comprises three parts: receiver-specific state reconstruction and graph construction; HV-intent prediction and a graph-attention decision policy; and timestamped CAV-intent exchange followed by a separate local conflict-aware arbitration layer. The following subsections introduce these parts in turn.

\subsection{Components of the Model}

\subsubsection{Receiver-Specific State Reconstruction and Graph Construction}

As shown in Fig.~\ref{fig:sign_mappo_framework}(a), each CAV
reconstructs a traffic-state view from the vehicle-state
information available to it. Delay and packet loss may cause
different CAVs to hold different states for the same vehicle
at a given decision step. The reconstructed state of vehicle
$j$ available to CAV $i$ is therefore denoted by
$\widetilde{\vect s}_{j\mid i,t}$.

To retain the vehicles most relevant to the current
intersection interaction while limiting graph size, CAV $i$
selects up to three vehicles nearest to the stop line on each
incoming lane and constructs its receiver-specific graph from
these vehicles and their relations:
\begin{equation}
G_{i,t}
=
\bigl(
\mathcal V_{i,t},
\mathcal E_{i,t},
\vect R_{i,t}
\bigr),
\label{eq:receiver_specific_graph}
\end{equation}
where $\mathcal V_{i,t}$ is the node set, $\mathcal E_{i,t}$ is the edge set, and $\vect R_{i,t}$ contains the edge features. Each selected vehicle corresponds to one node. Its ten-dimensional base feature is
\begin{equation}
\begin{aligned}
\vect x^{\mathrm{base}}_{j\mid i,t}
={}&
[\bar d,\bar v,\eta^{\mathrm{head}},\bar w,\bar q,
P,\bar o^{\mathrm{conf}},u^{\mathrm{turn}},
c,\bar a^{\mathrm{acc}}]^{\mathsf T}_{j\mid i,t}
\in\mathbb R^{10},
\end{aligned}
\label{eq:receiver_node_feature}
\end{equation}
where the common subscript $j\mid i,t$ applies to every component.
In order, the entries represent normalized stop-line distance
and speed, a lane-head indicator, normalized accumulated
waiting time and approach vehicle count, local priority,
normalized conflict-time overlap, turning and CAV indicators,
and normalized acceleration. The lane-head indicator identifies
the vehicle closest to the stop line on its lane in CAV $i$'s
observation. The turning and CAV indicators distinguish
left-turn from through routes and CAVs from HVs, respectively.
The approach count includes vehicles from the same approach
represented within the control region; continuous and
count-valued entries use fixed normalization ranges.

The local priority combines waiting time, approach traffic,
the relative traffic demand of conflicting approaches, and
stop-line proximity:
\begin{equation}
\begin{aligned}
\bar q^{\mathrm{adv}}_{j\mid i,t}
&=
\frac{1}{2}
\left[
1+
\frac{
q_{j\mid i,t}
-q^{\mathrm{conf,max}}_{j\mid i,t}
}{
\max\!\left(
q_{j\mid i,t}
+q^{\mathrm{conf,max}}_{j\mid i,t},
1
\right)}
\right],
\\
P_{j\mid i,t}
&=
\alpha_w\bar w_{j\mid i,t}
+\alpha_q\bar q_{j\mid i,t}
+\alpha_{\mathrm{adv}}\bar q^{\mathrm{adv}}_{j\mid i,t}
+\alpha_d(1-\bar d_{j\mid i,t}),
\end{aligned}
\label{eq:receiver_local_priority}
\end{equation}
where $q^{\mathrm{conf,max}}_{j\mid i,t}$ is the largest
receiver-observed vehicle count among approaches conflicting
with vehicle $j$'s route. The term
$\bar q^{\mathrm{adv}}_{j\mid i,t}$ compares this count with
the count on vehicle $j$'s approach. The coefficients
$\alpha_w$, $\alpha_q$, $\alpha_{\mathrm{adv}}$, and
$\alpha_d$ weight the four priority components, respectively.

Two vehicles are route-conflicting when their predefined
routes intersect within the conflict area. Their predicted
occupancy intervals in the pairwise conflict region are
obtained by projecting receiver-visible motion along the
routes under fixed acceleration and speed limits. The maximum
temporal overlap of vehicle $j$ is
\begin{equation}
o^{\mathrm{conf}}_{j\mid i,t}
=
\max_{k\in\mathcal C_{j\mid i,t}}
\Bigl[
\min\!\left(t^{\mathrm{out}}_{j,k\mid i,t},
t^{\mathrm{out}}_{k,j\mid i,t}\right)
-
\max\!\left(t^{\mathrm{in}}_{j,k\mid i,t},
t^{\mathrm{in}}_{k,j\mid i,t}\right)
\Bigr]_{+},
\label{eq:maximum_conflict_overlap}
\end{equation}
where $\mathcal C_{j\mid i,t}$ contains vehicles whose routes
conflict with that of vehicle $j$, and $t^{\mathrm{in}}$ and
$t^{\mathrm{out}}$ denote entry and exit times for the
corresponding pairwise conflict region. Here
$[x]_{+}=\max(x,0)$, and the overlap is set to zero when
$\mathcal C_{j\mid i,t}$ is empty. Because these times are
estimated from receiver-visible states, overlap features may
differ across CAVs.

The graph includes bidirectional route-conflict edges between
selected vehicles with intersecting routes and directed
same-lane-following edges from the nearest follower to the
lane-head vehicle. For each edge $(u,v)\in\mathcal E_{i,t}$, its feature is defined as
\begin{equation}
\vect r_{uv\mid i,t}
=
\left[
\min\!\left(
\frac{
\left\|
\vect p_{u\mid i,t}-\vect p_{v\mid i,t}
\right\|_2
}{100},
1
\right),
\eta^{\mathrm{rel}}_{uv\mid i,t}
\right]^{\mathsf T}
\in\mathbb R^2,
\label{eq:receiver_edge_feature}
\end{equation}
where $\vect p_{u\mid i,t}$ and $\vect p_{v\mid i,t}$ are the
receiver-visible planar positions of the source and target
vehicles. The first entry is their distance divided by
100~m and capped at $1$; the edge-type indicator
$\eta^{\mathrm{rel}}_{uv\mid i,t}$ equals $0$ for a
route-conflict edge and $1$ for a same-lane-following edge.

\subsubsection{HV Intent Prediction and Graph-Attention Policy}

Because HVs do not communicate their crossing intentions,
SIGN-MAPPO predicts HV-intent features from the motion
histories available to each CAV and incorporates them into
the graph-attention policy, as shown in
Fig.~\ref{fig:sign_mappo_framework}(b). For HV $h$ in the
graph of CAV $i$, the predictor uses
\begin{equation}
\vect H^{\mathrm{HV}}_{h\mid i,t}
=
\left[
\vect x^{\mathrm{hist}}_{h\mid i,t-n+1},\ldots,
\vect x^{\mathrm{hist}}_{h\mid i,t}
\right]^{\mathsf T}
\in\mathbb R^{n\times10}.
\label{eq:hv_intent_history}
\end{equation}
Each row contains speed, acceleration, jerk, stop-line distance, accumulated waiting time, speed variation, the number of stop--go transitions, high-acceleration events, hesitation events, and a long-wait indicator. A shared gated recurrent unit (GRU) processes these rows in temporal order:
\begin{equation}
\begin{aligned}
\vect g^{\mathrm{HV}}_{h\mid i,\tau}
&=
\operatorname{GRU}_{\psi}\!\left(
\vect x^{\mathrm{hist}}_{h\mid i,\tau},
\vect g^{\mathrm{HV}}_{h\mid i,\tau-1}
\right),\\
\vect z_{h\mid i,t}
&=
f_{z,\psi}\!\left(
\vect g^{\mathrm{HV}}_{h\mid i,t}
\right)
\in\mathbb R^6,
\end{aligned}
\label{eq:hv_intent_encoding}
\end{equation}
where $\vect g^{\mathrm{HV}}_{h\mid i,\tau}$ retains the information accumulated up to step $\tau$, and $\vect z_{h\mid i,t}$ is the resulting latent representation of the HV's recent motion.

The latent representation is passed to three sigmoid output heads:
\begin{equation}
\left[
p^{\mathrm{go}}_{h\mid i,t},
p^{\mathrm{enter}}_{h\mid i,t},
p^{\mathrm{aggr}}_{h\mid i,t}
\right]^{\mathsf T}
=
\sigma\!\left(f_{p,\psi}(\vect z_{h\mid i,t})\right).
\label{eq:hv_intent_outputs}
\end{equation}
They represent movement tendency, short-horizon conflict-area
entry risk, and aggressive-type probability, respectively.
The prediction horizon for conflict-area entry is $1$~s.

The predictor output is converted into the six-dimensional feature supplied to the graph policy:
\begin{equation}
\vect e^{\mathrm{HV}}_{h\mid i,t}
=
\left[
z_1,z_2,z_3,
p^{\mathrm{go}},p^{\mathrm{enter}},p^{\mathrm{aggr}}
\right]^{\mathsf T}_{h\mid i,t}
\in\mathbb R^6,
\label{eq:hv_intent_feature}
\end{equation}
where the first three entries retain latent motion information, and the final three entries are the intent outputs defined in (\ref{eq:hv_intent_outputs}). The remaining coordinates of $\vect z_{h\mid i,t}$ are used by the prediction heads but are not separately included in $\vect e^{\mathrm{HV}}_{h\mid i,t}$. When sufficient history is available, a reliability-weighted copy of this feature is placed in the row of the corresponding HV. CAV rows and HV rows without an available prediction receive six-dimensional zero vectors. Stacking these rows gives $\vect E^{\mathrm{HV}}_{i,t}$, and the input node matrix becomes
\begin{equation}
\vect H^{(0)}_{i,t}
=
\left[
\vect X^{\mathrm{base}}_{i,t}
\Vert
\vect E^{\mathrm{HV}}_{i,t}
\right]
\in\mathbb R^{N_{i,t}\times16}.
\label{eq:gat_input_feature}
\end{equation}

The shared GAT actor processes the augmented node features
together with the graph edges and their two-dimensional
features. An input projection and two residual GAT layers
update the node representations:
\begin{equation}
\begin{aligned}
\vect Z^{(0)}_{i,t}
&=
\operatorname{ReLU}\!\left(
\vect H^{(0)}_{i,t}\vect W_{\mathrm{in}}
+\vect b_{\mathrm{in}}
\right),\\
\vect Z^{(\ell)}_{i,t}
&=
\operatorname{LN}\!\left[
\vect Z^{(\ell-1)}_{i,t}
+\operatorname{GAT}^{(\ell)}_{\theta}\!\left(
\vect Z^{(\ell-1)}_{i,t},
\mathcal E_{i,t},
\{\vect r_{uv\mid i,t}\}
\right)
\right],
\end{aligned}
\label{eq:gat_policy_layers}
\end{equation}
where $\vect W_{\mathrm{in}}$ and $\vect b_{\mathrm{in}}$ project each 16-dimensional node feature into the initial hidden representation $\vect Z^{(0)}_{i,t}$. At layer $\ell$, $\operatorname{GAT}^{(\ell)}_{\theta}$ aggregates the representations of nodes connected by $\mathcal E_{i,t}$ and uses $\vect r_{uv\mid i,t}$ when calculating their attention weights. The aggregated result is added to the preceding representation through the residual connection, and $\operatorname{LN}(\cdot)$ denotes layer normalization. The policy head then maps the updated node representations to action logits:
\begin{equation}
\vect L_{i,t}
=
f_{\pi,\theta}\!\left(\vect Z^{(2)}_{i,t}\right)
\in\mathbb R^{N_{i,t}\times2},
\qquad
a^{\mathrm{raw}}_{i,t}
\sim
\pi_{\theta}(\cdot\mid G_{i,t},i),
\label{eq:graph_policy_output}
\end{equation}
where $f_{\pi,\theta}$ maps the final GAT representations to
the logit matrix $\vect L_{i,t}$. For CAV $i$, the two values
in its corresponding row are denoted by
$l^{\mathrm{Stop}}_{i,t}$ and $l^{\mathrm{Go}}_{i,t}$, representing
its preferences for \emph{Stop} and \emph{Go}, respectively.
CAV $i$
uses the resulting action distribution
$\pi_{\theta}(\cdot\mid G_{i,t},i)$ to generate its raw
crossing decision $a^{\mathrm{raw}}_{i,t}$.

\subsubsection{Timestamped CAV-Intent Exchange and Local Conflict-Aware Arbitration}

The graph policy generates a raw crossing intention for each
decision-eligible CAV independently. Simultaneous \emph{Go}
intentions from route-conflicting CAVs may be incompatible.
As shown in Fig.~\ref{fig:sign_mappo_framework}(c), CAVs
therefore exchange their intentions before applying local
conflict-aware arbitration.

After generating its raw action, CAV $j$ forms the timestamped intent message
\begin{equation}
\mu^{\mathrm{tx}}_{j,t}
=
\left(j,a^{\mathrm{raw}}_{j,t},t\right),
\label{eq:cav_intent_message}
\end{equation}
where $j$ identifies the sending CAV,
$a^{\mathrm{raw}}_{j,t}$ is its raw \emph{Stop}/\emph{Go}
intention, and $t$ is the decision-step timestamp at which
the intention is generated. The message is transmitted to the other decision-eligible CAVs through V2V communication. Communication delay and packet loss determine when, or whether, the message becomes available to each receiver.

Each CAV maintains a separate intent cache for every sender. For receiver $i$, the cached record of sender $j$ is
\begin{equation}
\widetilde\mu_{j\mid i,t}
=
\left(
j,
a^{\mathrm{raw}}_{j,\tau^{\mathrm{int}}_{j\rightarrow i}(t)},
\tau^{\mathrm{int}}_{j\rightarrow i}(t)
\right),
\label{eq:cav_intent_cache}
\end{equation}
where $\tau^{\mathrm{int}}_{j\rightarrow i}(t)\le t$ is the
decision-step timestamp of the latest intent message from
CAV $j$ received by CAV $i$. If no new message is delivered, the preceding valid record remains in the cache. The intent records available to CAV $i$ are collected in
\begin{equation}
\mathcal M_{i,t}
=
\left\{
\widetilde\mu_{j\mid i,t}
\right\}.
\label{eq:cav_intent_set}
\end{equation}
Because the sender--receiver links experience delay and packet loss independently, the contents of $\mathcal M_{i,t}$ may differ among receiving CAVs.

A raw \emph{Stop} intention remains unchanged. For a raw
\emph{Go} intention, CAV $i$ considers its own request and
the received \emph{Go} requests from other CAVs. It calculates
their priorities from its receiver-specific state information
using (\ref{eq:receiver_local_priority}) and examines the
requests in descending priority order. A \emph{Go} request is
retained if its route and estimated conflict occupancy do not
conflict with higher-priority requests already retained;
otherwise, it is changed to \emph{Stop}. Thus, CAVs with
compatible movements may retain \emph{Go} at the same
decision step.

CAV-intent arbitration alone cannot account for conflicts with
HVs because HVs do not transmit their intentions and may enter
the conflict area while a CAV intends to proceed. The arbitration procedure also includes a kinematics-based
HV-conflict check. An HV is considered when it has entered
the conflict area, or when it is within the screening distance
and its speed or acceleration exceeds the corresponding
threshold. If its route conflicts with that of CAV $i$ and
their estimated occupancy intervals overlap, CAV $i$'s
\emph{Go} intention is changed to \emph{Stop}.

To prevent prolonged mutual waiting, a progress check is
applied when all candidate \emph{Go} requests have been
rejected and persistent waiting or queue accumulation is
detected. The rejected candidates are examined again, and
one that passes the conflict-overlap and HV-risk checks may
be permitted to \emph{Go}.

Every decision-eligible CAV applies the above procedure independently using the state and intent information available to itself. Its executable action is written as
\begin{equation}
a^{\mathrm{exe}}_{i,t}
=
\mathcal F_i^{\mathrm{arb}}\!\left(
a^{\mathrm{raw}}_{i,t},
\mathcal M_{i,t},
\vect o_{i,t}
\right),
\label{eq:local_arbitration}
\end{equation}
where $\mathcal F_i^{\mathrm{arb}}$ denotes the receiver-side arbitration function. The resulting $a^{\mathrm{exe}}_{i,t}$ is the action applied by CAV $i$ at decision step $t$.

\subsubsection{Shared-Parameter Multi-Agent PPO Training and Receiver-Specific Execution}

\begin{figure}[!t]
\centering
\includegraphics[width=\columnwidth]{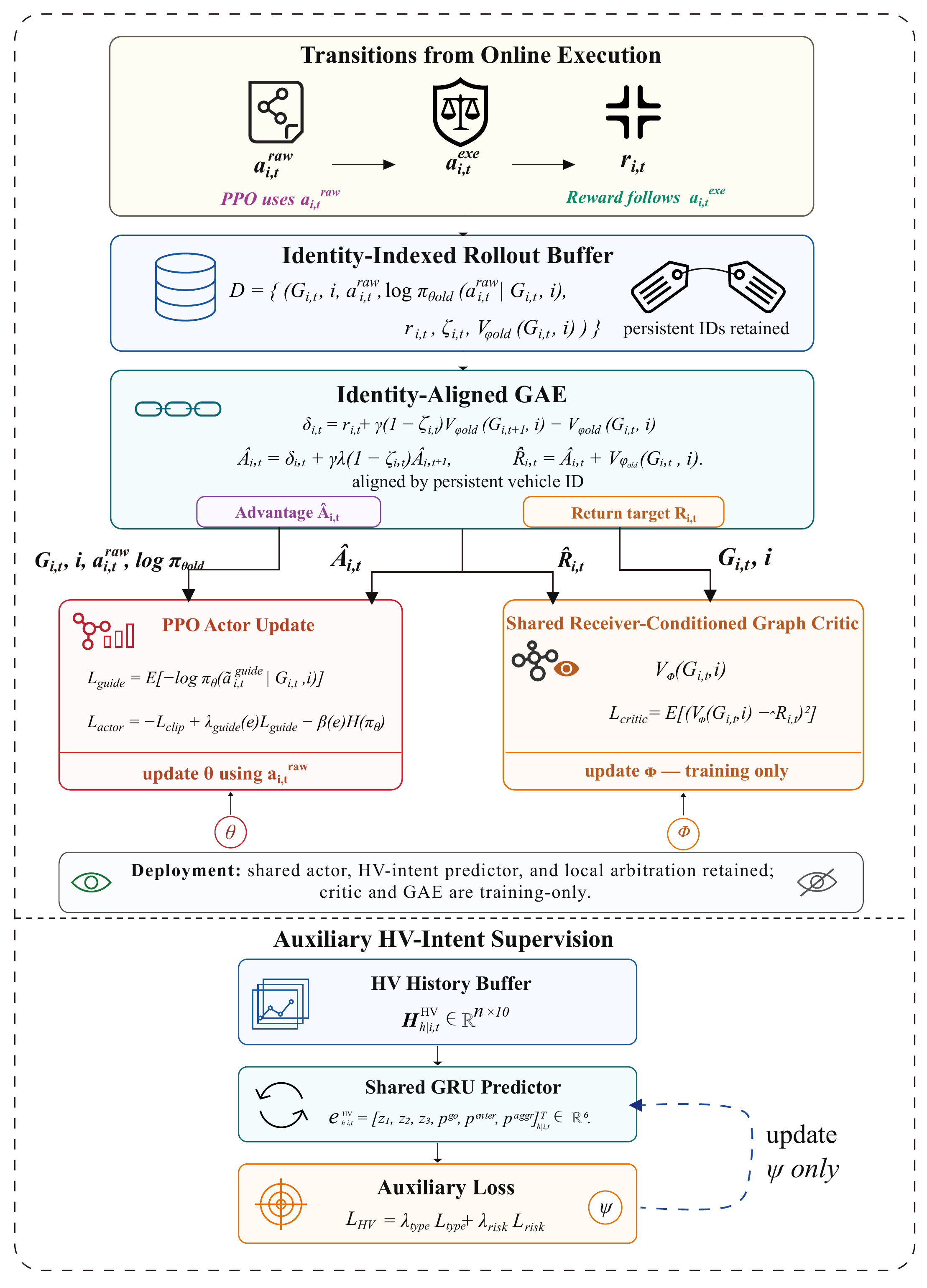}
\caption{Shared-parameter multi-agent PPO training and receiver-specific execution in SIGN-MAPPO.}
\label{fig:ctde_training}
\end{figure}

As illustrated in Fig.~\ref{fig:ctde_training}, SIGN-MAPPO
shares actor and critic parameters across decision-eligible
CAVs while retaining a separate graph $G_{i,t}$ for each
receiver. Trajectories from all CAVs are pooled for MAPPO
updates \cite{yu2022mappo}, whereas each CAV makes its
execution-time decisions using the information available to
itself.

The actor $\pi_{\theta}$ and critic $V_{\phi}$ are shared by all CAVs. For CAV $i$, both networks process $G_{i,t}$ and read the representation corresponding to that CAV. The actor generates $a^{\mathrm{raw}}_{i,t}$, and the critic estimates
\begin{equation}
V_{i,t}=V_{\phi}(G_{i,t},i),
\label{eq:receiver_conditioned_value}
\end{equation}
where $V_{i,t}$ is the expected return associated with the decision of CAV $i$. The parameters $\theta$ and $\phi$ are common to all CAVs, but the network outputs vary with their receiver-specific graph inputs. Centralized training refers to pooling these CAV trajectories and updating the shared networks together. The critic is evaluated separately for each CAV on its receiver-specific graph.

The decision-eligible set changes as vehicles enter and leave
the intersection, so adjacent samples cannot be linked by
their positions in a batch. Each transition is stored with
the persistent CAV identity, receiver-specific graph, raw
action, old-policy log probability, reward, termination flag,
and critic estimate. Samples are aligned by identity before
advantages and returns are calculated, preventing the
trajectory of one CAV from being continued with another
CAV's value estimate.

For two consecutive samples belonging to CAV $i$, the temporal-difference residual, generalized advantage estimate, and return target are
\begin{align}
\delta_{i,t}
&=
r_{i,t}
+\gamma(1-\zeta_{i,t})
V_{\phi_{\mathrm{old}}}(G_{i,t+1},i)
-V_{\phi_{\mathrm{old}}}(G_{i,t},i),
\notag\\
\widehat A_{i,t}
&=
\delta_{i,t}
+\gamma\lambda(1-\zeta_{i,t})\widehat A_{i,t+1},
\notag\\
\widehat R_{i,t}
&=
\widehat A_{i,t}
+V_{\phi_{\mathrm{old}}}(G_{i,t},i),
\label{eq:identity_aligned_gae}
\end{align}
where $r_{i,t}$ is CAV $i$'s reward, $\gamma$ is the
discount factor, and $\lambda$ is the GAE trace-decay factor.
$V_{\phi_{\mathrm{old}}}$ denotes the critic before the
current update; $\delta_{i,t}$, $\widehat A_{i,t}$, and
$\widehat R_{i,t}$ are the temporal-difference residual,
advantage estimate, and critic return target, respectively.
The flag $\zeta_{i,t}$ equals $1$ when CAV $i$'s trajectory
terminates or has no valid subsequent sample, and $0$
otherwise. When $\zeta_{i,t}=0$, $G_{i,t+1}$ is the next
receiver-specific graph of the same CAV.

The shared actor and critic are optimized using the standard
MAPPO objective \cite{yu2022mappo}, with two adaptations to
rollout attribution. First, The PPO probability ratio is computed using the raw action
$a^{\mathrm{raw}}_{i,t}$ sampled by the actor. The reward
$r_{i,t}$ and the next receiver-specific graph
$\mathcal G_{i,t+1}$ are obtained after the corresponding
executable action $a^{\mathrm{exe}}_{i,t}$ is applied to the
traffic environment. The policy update is therefore associated with the
action sampled from $\pi_{\theta}$, while its return reflects
the traffic outcome produced by arbitration-adjusted execution.
Second, the advantage estimates and return targets are computed
from the identity-aligned trajectories in
(\ref{eq:identity_aligned_gae}), ensuring that consecutive
samples belong to the same CAV when the decision-eligible set
changes. The standard MAPPO clipped surrogate objective is
denoted by $\mathcal L_{\mathrm{clip}}$ in the following actor
objective.

At the beginning of training, the raw actions produced by the actor may frequently be changed by arbitration. A guidance target is therefore obtained from the priority and conflict checks and used to help the actor learn decisions that are more consistent with the coordination result. The guidance loss is
\begin{equation}
\mathcal L_{\mathrm{guide}}
=
\mathbb E\!\left[
-\log\pi_{\theta}\!\left(
\widetilde a^{\mathrm{guide}}_{i,t}
\mid G_{i,t},i
\right)
\right],
\label{eq:arbitration_guidance_loss}
\end{equation}
where $\widetilde a^{\mathrm{guide}}_{i,t}$ is the action recommended by the arbitration conditions for the current training sample. Its influence is reduced during training so that the final policy is primarily optimized by the interaction return. The actor and critic objectives are
\begin{align}
\mathcal L_{\mathrm{actor}}
&=
-\mathcal L_{\mathrm{clip}}
+\lambda_{\mathrm{guide}}(e)\mathcal L_{\mathrm{guide}}
-\beta(e)\mathcal H(\pi_{\theta}),
\label{eq:actor_loss}\\
\mathcal L_{\mathrm{critic}}
&=
\mathbb E\!\left[
\left(
V_{\phi}(G_{i,t},i)-\widehat R_{i,t}
\right)^2
\right],
\label{eq:critic_loss}
\end{align}
where $e$ is the training-episode index, $\lambda_{\mathrm{guide}}(e)$ controls the guidance loss, $\beta(e)$ controls entropy regularization, and $\mathcal H(\pi_{\theta})$ denotes the policy entropy.

The environmental reward does not directly supervise the meaning of the three HV-intent outputs. The HV-intent predictor is therefore updated with the auxiliary objective
\begin{equation}
\begin{aligned}
\mathcal L_{\mathrm{HV}}
&=
\lambda_{\mathrm{type}}\mathcal L_{\mathrm{type}}
+\lambda_{\mathrm{risk}}\mathcal L_{\mathrm{risk}},\\
\mathcal L_{\mathrm{risk}}
&=
\frac{1}{3}
\sum_{r\in\{\mathrm{go},\mathrm{enter},\mathrm{aggr}\}}
\operatorname{BCE}\!\left(
p^r_{h\mid i,t},y^r_{h\mid i,t}
\right),
\end{aligned}
\label{eq:hv_intent_loss}
\end{equation}
where $\lambda_{\mathrm{type}}$ and
$\lambda_{\mathrm{risk}}$ weight the two loss terms, and
$\mathcal L_{\mathrm{type}}$ is the classification loss for
aggressive and non-aggressive HVs. The target
$y^r_{h\mid i,t}$ is the binary label of intent output $r$, and
$\operatorname{BCE}(\cdot)$ denotes binary cross-entropy. The
predictor parameters $\psi$ are updated using the HV histories
collected during rollout.

During online execution, each CAV retains receiver-specific
state reconstruction, HV-intent prediction, the shared actor,
CAV-intent exchange, and local arbitration. The critic,
advantage estimation, and parameter updates are used only
during training.

\subsection{Implementation of the Algorithm}

At each decision step, each eligible CAV reconstructs its
receiver-specific traffic state, builds the corresponding graph,
and augments HV nodes with predicted intent features. The
shared GAT actor generates a raw CAV intention, which is
exchanged through the timestamped intent channel and
converted into an executable action by local arbitration.

During training, identity-indexed transitions are collected
to compute aligned advantages and returns for updating the
shared actor and critic, while HV histories supervise the
intent predictor. During online execution, only state
reconstruction, HV-intent prediction, the GAT actor,
CAV-intent exchange, and local arbitration are retained.
\textbf{Algorithm~1} summarizes the complete procedure.

\begin{figure}[!t]
\refstepcounter{algorithm}
\noindent\rule{\columnwidth}{0.8pt}
\noindent\textbf{Algorithm \thealgorithm:} Training and Online Execution of SIGN-MAPPO.\par
\vspace{1pt}
\noindent\rule{\columnwidth}{0.4pt}
\label{alg:sign_mappo}
\footnotesize
\setlength{\tabcolsep}{0pt}
\renewcommand{\arraystretch}{1.03}
\begin{tabularx}{\columnwidth}{@{}r@{\hspace{0.5em}}X@{}}
1 & \textbf{Input:} shared actor parameters $\theta$, critic parameters $\phi$, predictor parameters $\psi$, and rollout horizon $T$.\\
2 & \textbf{for each training iteration do}\\
3 & \quad Reset the traffic environment and receiver-side communication caches.\\
4 & \quad \textbf{for} $t=0,\ldots,T-1$ \textbf{do}\\
5 & \qquad Determine the decision-eligible CAV set $\mathcal I_t$.\\
6 & \qquad \textbf{for each} $i\in\mathcal I_t$ \textbf{do}\\
7 & \qquad\quad Reconstruct the traffic-state observation $\vect o_{i,t}$.\\
8 & \qquad\quad Construct $G_{i,t}$ with its node and edge features.\\
9 & \qquad\quad Infer the HV-intent features and form $\vect H^{(0)}_{i,t}$.\\
10 & \qquad\quad Apply the GAT actor and obtain $\vect L_{i,t}$.\\
11 & \qquad\quad Sample the raw intention $a^{\mathrm{raw}}_{i,t}$.\\
12 & \qquad \textbf{end for}\\
13 & \qquad Transmit the timestamped CAV-intent messages.\\
14 & \qquad Update each CAV's intent cache with the delivered messages.\\
15 & \qquad \textbf{for each} $i\in\mathcal I_t$ \textbf{do}\\
16 & \qquad\quad Apply $\mathcal F_i^{\mathrm{arb}}$ using $a^{\mathrm{raw}}_{i,t}$, $\mathcal M_{i,t}$, and $\vect o_{i,t}$.\\
17 & \qquad\quad Obtain the executable action $a^{\mathrm{exe}}_{i,t}$.\\
18 & \qquad \textbf{end for}\\
19 & \qquad Execute $\{a^{\mathrm{exe}}_{i,t}\}_{i\in\mathcal I_t}$ in the traffic environment.\\
20 & \qquad Store the identity-indexed transition for each CAV.\\
21 & \quad \textbf{end for}\\
22 & \quad Align the rollout samples by CAV identity.\\
23 & \quad Compute $\widehat A_{i,t}$ and $\widehat R_{i,t}$ using identity-aligned GAE.\\
24 & \quad Update $\theta$ using $\mathcal L_{\mathrm{actor}}$ and update $\phi$ using $\mathcal L_{\mathrm{critic}}$.\\
25 & \quad Update $\psi$ using $\mathcal L_{\mathrm{HV}}$.\\
26 & \textbf{end for}\\
27 & \textbf{Online execution:} Repeat Steps 3--21 without Step 20, and omit Steps 22--25.\\
\end{tabularx}
\vspace{1pt}
\noindent\rule{\columnwidth}{0.8pt}
\end{figure}

\section{Experimental Evaluation}
\label{sec:experiments}

The experimental evaluation is organized around three
objectives. First, the training curves are used to assess whether
SIGN-MAPPO converges to a stable decision policy. The
comparison with other controllers and the All-Go diagnostic
case then evaluates its relative performance in terms of
completion rate, collision rate, and mean intersection traversal
time. Second, the module ablations examine the respective
roles of CAV-intent exchange and HV-intent prediction, while
the predictor evaluation measures the accuracy of the inferred
HV intent. Third, the state-channel delay and packet-loss
robustness experiments and the joint state--intent delay and
packet-loss robustness experiments characterize the
communication robustness of the trained SIGN-MAPPO policy,
while the cross-scenario evaluation examines its transferability
across different road, traffic, and HV-behavior settings.

\subsection{Experiments Setup}
\label{subsec:experiments_Setup}

\subsubsection{Simulation and Parameter Settings}

To evaluate SIGN-MAPPO under repeatable mixed-traffic
settings, the experiments are conducted using Simulation of
Urban MObility (SUMO)~\cite{lopez2018sumo}. SUMO generates
the traffic flow and HV motion, updates the vehicle states, and
applies the executable CAV actions produced by each evaluated
controller.

The principal scenario follows the four-leg unsignalized
intersection defined in Section~III, with one left-turn lane
and two through lanes on each approach. It contains 90
vehicles under high demand, with a CAV penetration rate of
50\% and equal proportions of hesitant, nominal, and aggressive
HVs. The target speed is 13.89~m/s (50~km/h), and the control region extends 55~m upstream
of each stop line. At the target
speed, traversing this region takes approximately 4~s. With
a decision interval of 0.1~s, the CAV can update its available
state and intent information and revise its crossing decision
approximately 40 times before reaching the stop line.

The simulation and evaluation settings are summarized in
Table~\ref{tab:simulation_settings}, and the model and training
parameters are listed in Table~\ref{tab:training_settings}.
All evaluations are conducted with the trained parameters
held fixed.

\begin{table}[!t]
\caption{Principal Simulation and Evaluation Settings}
\label{tab:simulation_settings}
\centering
\footnotesize
\begin{tabularx}{\columnwidth}{@{}p{0.37\columnwidth}X@{}}
\toprule
\textbf{Item} & \textbf{Setting}\\
\midrule
Intersection & Four-leg unsignalized intersection\\
Inbound lanes & One left-turn and two through lanes per approach\\
Control-region length & 55~m upstream of each stop line\\
Simulation step / episode limit & 0.1~s / 6000 steps\\
Main demand & 90 arriving vehicles\\
CAV penetration & 50\%\\
HV composition & Hesitant/nominal/aggressive = 1:1:1\\
Action space & \emph{Stop}/\emph{Go} for lane-head CAVs\\
Formal evaluation & Seeds 2026--2028; 15 episodes per seed\\
Evaluation episodes & 45 (15 per seed; seeds 2026--2028)\\
Figure error bars & 95\% CIs\\
\bottomrule
\end{tabularx}
\end{table}

\begin{table}[!t]
\caption{Principal Model and Training Settings}
\label{tab:training_settings}
\centering
\scriptsize
\begin{tabularx}{\columnwidth}{@{}p{0.45\columnwidth}X@{}}
\toprule
\textbf{Parameter} & \textbf{Value}\\
\midrule
GAT layers / heads / hidden size & 2 / 4 / 64\\
Base / HV-intent feature size & 10 / 6\\
Maximum HV history / GRU hidden size & $30\times10$ / 64\\
Actor / critic learning rate & $3\times10^{-4}$ / $10^{-3}$\\
Discount $\gamma$ / GAE $\lambda$ & 0.99 / 0.85\\
PPO clip / epochs / minibatch & 0.20 / 4 / 128\\
Rollout / maximum samples & 1000 / 3000\\
Entropy coefficient & Linear decay, 0.05 to 0.005\\
HV risk-loss coefficient & 0.6\\
Maximum HV-feature weight & 0.50\\
Training episodes & 300\\
Priority coefficients & 0.40 / 0.30 / 0.20 / 0.10\\
\bottomrule
\end{tabularx}
\end{table}

\subsubsection{Evaluation Metrics}
\par
\begin{itemize}
    \item \emph{Completion rate}: The percentage of generated
    vehicles that exit the conflict area within an episode.

    \item \emph{Collision rate}: The percentage of generated
    vehicles involved in a recorded collision.

    \item \emph{Mean intersection traversal time}: The average
    time from control-region entry to conflict-area exit for
    vehicles that complete the crossing.

    \item \emph{Waiting time}: The mean accumulated SUMO
    waiting time across CAV observations within the control
    region.

    \item \emph{Arbitration intervention rate}: The percentage
    of raw CAV intentions changed by the arbitration procedure.
\end{itemize}

\subsubsection{Compared Methods}

SIGN-MAPPO is compared with the following six baseline
approaches.

\emph{FCFS-Priority}~\cite{dresner2008multiagent}:
Vehicles are ordered upon entering the control region and
released accordingly; nonconflicting lane-head CAVs may
proceed simultaneously.

\emph{Conflict-TTC}~\cite{levin2017conflict,hayward1972near}:
This rule-based method combines route conflicts with
time-to-collision screening. It estimates arrival time as
$d/\max(v,v_{\min})$ and stops the lower-priority vehicle
when the arrival-time difference falls below the safety gap.

\emph{Mixed-Traffic Controller}~\cite{wang2025mixed}:
The public implementation by Wang \emph{et al.} is adapted
to the present intersection and retrained while retaining
its original model and learning procedure. 

\emph{MLP-MAPPO}~\cite{yu2022mappo}:
Node-wise multilayer perceptrons replace SIGN-MAPPO's graph
networks. The MAPPO rollout, reward, and update procedure
are retained, but graph relations and both intent modules
are removed.

\emph{Attn-MAPPO}~\cite{zhuang2023cooperative}:
This reimplementation applies the reported global
self-attention architecture to selected vehicle nodes,
without explicit graph relations or intent features.

\emph{All-Go}:
This initial baseline assigns \emph{Go} to every
decision-eligible CAV without a learned policy or conflict
arbitration.

\subsection{Training Performance}
\label{subsec:training_performance}

\begin{figure*}[!t]
\centering
\includegraphics[width=\textwidth]{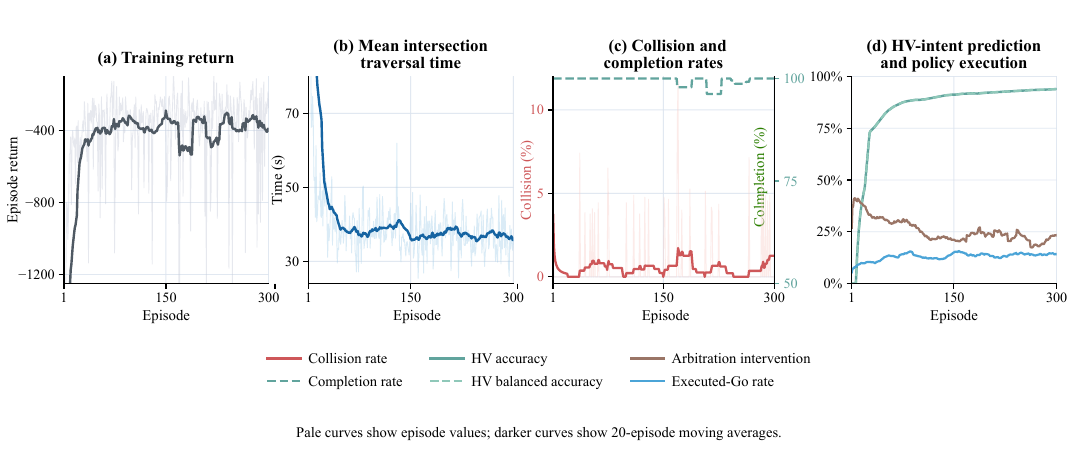}
\caption{Training dynamics of SIGN-MAPPO over 300 episodes.
Panels (a)--(c) show episode-level values and their
20-episode moving averages (MA20), while panel (d) reports
the MA20 values of the HV-intent prediction and policy-
execution indicators.}
\label{fig:training_curves}
\end{figure*}
The first evaluation objective begins by assessing whether
SIGN-MAPPO converges to stable decision performance.
As shown in Fig.~\ref{fig:training_curves}(a), the episode
return rises rapidly during initial exploration and then
fluctuates within a higher stable range. In
Fig.~\ref{fig:training_curves}(b), mean intersection traversal
time decreases from above 70~s to approximately 35--40~s
before stabilizing. Figure~\ref{fig:training_curves}(c) shows
that the 20-episode moving average of completion rate
approaches 100\%, while the collision-rate moving average
remains low despite occasional episode-level fluctuations.
Together, these trends indicate stable policy performance
in traffic completion, safety, and passage efficiency.

Figure~\ref{fig:training_curves}(d) presents the HV-intent
prediction and execution-side indicators. The overall and
balanced accuracies of the HV-intent predictor rise and
stabilize, indicating increasingly stable predicted features
for the graph policy. The arbitration intervention rate
decreases from approximately 40\% to 20--25\%, showing
that a growing proportion of raw CAV intentions are accepted
without modification. The executed-\emph{Go} rate stabilizes
at approximately 12--15\%, indicating a stable final action
pattern.

\FloatBarrier

\subsection{Comparative Experiments}
\label{subsec:comparative_experiments}

Following the convergence assessment, the first evaluation objective compares SIGN-MAPPO with six baselines, including All-Go, in terms of traffic completion, safety, and passage efficiency. All are evaluated under the same
intersection layout, traffic demand, vehicle composition,
episode length, and seeds, with method-specific architectures
and training settings retained as described above. Table~
\ref{tab:main_comparison} reports the results.

\begin{table}[!t]
\caption{Main Comparison Over 45 Test Episodes}
\label{tab:main_comparison}
\centering
\scriptsize
\setlength{\tabcolsep}{2pt}
\renewcommand{\arraystretch}{1.05}
\begin{tabularx}{\columnwidth}{
@{}>{\raggedright\arraybackslash}X
*{3}{>{\centering\arraybackslash}p{0.42in}}@{}}
\toprule
\textbf{Method} & \textbf{Comp.} & \textbf{Coll.} & \textbf{Trav.}\\
& \textbf{(\%) $\uparrow$}
& \textbf{(\%) $\downarrow$}
& \textbf{(s) $\downarrow$}\\
\midrule
Mixed-Traffic Controller~\cite{wang2025mixed}
& 92.56 & \textbf{0.00} & 52.89\\
Conflict-TTC~\cite{levin2017conflict,hayward1972near}
& \textbf{100.00} & 0.20 & 97.47\\
FCFS-Priority~\cite{dresner2008multiagent}
& 92.30 & 1.17 & 58.18\\
MLP-MAPPO~\cite{yu2022mappo}
& 74.64 & 3.89 & 31.13\\
Attn-MAPPO~\cite{zhuang2023cooperative}
& 73.04 & 6.68 & \textbf{30.00}\\
All-Go
& 70.02 & 6.56 & 30.37\\
\midrule
\textbf{SIGN-MAPPO}
& 97.53 & 0.10 & 35.28\\
\bottomrule
\end{tabularx}
\end{table}

For traffic completion, Conflict-TTC clears all generated
vehicles, while SIGN-MAPPO achieves 97.53\%. The complete
throughput of Conflict-TTC comes with a much longer traversal
time, indicating that its release decisions are comparatively
conservative. The Mixed-Traffic Controller and FCFS-Priority
complete approximately 92\% of vehicles, whereas MLP-MAPPO
and Attn-MAPPO remain below 75\%. SIGN-MAPPO's high completion
rate is consistent with its progress reward and queue penalties
discouraging prolonged accumulation, while CAV-intent exchange
and local arbitration help resolve competing lane-head requests.
The two learning baselines, which do not use the same explicit
route relations and separate intent pathways, are less effective
at clearing queues within the episode limit.

The Mixed-Traffic Controller records no collisions, but this
result is accompanied by lower completion and longer traversal
time than SIGN-MAPPO. SIGN-MAPPO records a 0.10\% collision
rate, followed by Conflict-TTC at 0.20\%; FCFS-Priority and
the two learning baselines have higher rates. The low collision
rate of SIGN-MAPPO is consistent with route-conflict relations
identifying incompatible movements, HV-intent features
informing the policy under uncertain HV motion, and local
arbitration checking raw CAV intentions before execution.
Unlike the zero-collision Mixed-Traffic Controller, it achieves
this safety level without a comparable reduction in completion
or passage efficiency.

The mean intersection traversal time of SIGN-MAPPO is
35.28~s, substantially shorter than the 52.89~s of the
Mixed-Traffic Controller, 58.18~s of FCFS-Priority, and
97.47~s of Conflict-TTC. Its route-conflict representation
allows compatible movements to proceed concurrently, reducing
unnecessary sequential release. MLP-MAPPO and Attn-MAPPO
report shorter times of 31.13 and 30.00~s, respectively.
However, these means are calculated only for the fewer than
75\% of vehicles that complete the crossing; they cannot be
interpreted as higher overall passage efficiency without
considering the uncompleted vehicles.

The initial All-Go baseline releases CAVs without coordinating
conflicting crossing requests. It completes 70.02\% of
vehicles and records a 6.56\% collision rate, compared with
97.53\% and 0.10\% for SIGN-MAPPO. Its mean traversal time
is shorter (30.37 versus 35.28~s), but this value includes
only the vehicles that complete the crossing. Thus, the
shorter recorded time does not offset the substantially lower
completion and safety performance of unconditional release.

Within the evaluated setting, SIGN-MAPPO does not lead on
each metric individually. It nevertheless combines high
completion and low collision exposure with considerably
shorter traversal times than the more conservative rule-based
approaches, providing a more balanced outcome across the
three measures.

\FloatBarrier

\subsection{Ablation Experiments}
\label{subsec:ablations}

The second evaluation objective examines the respective
contributions of CAV-intent exchange and HV-intent prediction.
The two modules are removed separately while the remaining
policy and execution mechanisms are retained. Direct predictor
evaluation is subsequently conducted to examine the quality of
the inferred HV-intent information.

\begin{figure}[!t]
\centering
\includegraphics[width=\columnwidth]{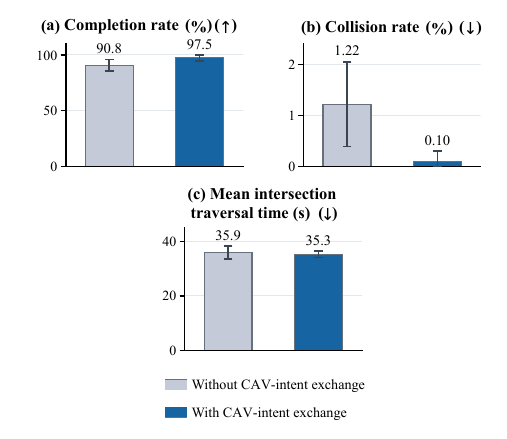}
\caption{CAV-intent exchange ablation. Error bars denote
95\% CIs.}
\label{fig:cav_intent_ablation}
\end{figure}

The CAV-intent ablation retains the receiver-specific state
graph and HV-intent predictor but removes the exchange of
timestamped raw CAV intentions. The experiment isolates the
contribution of CAV-intent exchange to coordination among
lane-head CAVs.

As shown in Fig.~\ref{fig:cav_intent_ablation}, removing
CAV-intent exchange reduces completion from 97.53\% to
90.77\% and increases collisions from 0.10\% to 1.22\%,
while mean intersection traversal time changes only from
35.28 to 35.85~s. The ablation thus affects traffic clearance
and safety more than the traversal time of completed vehicles.
In SIGN-MAPPO, exchanged intentions allow receiver-side
arbitration to identify competing \emph{Go} requests and
retain compatible movements. Without this coordination,
conflicting releases become more likely, and unresolved
competition among lane-head CAVs leaves more vehicles
unfinished.

\begin{figure}[!t]
\centering
\includegraphics[width=\columnwidth]
{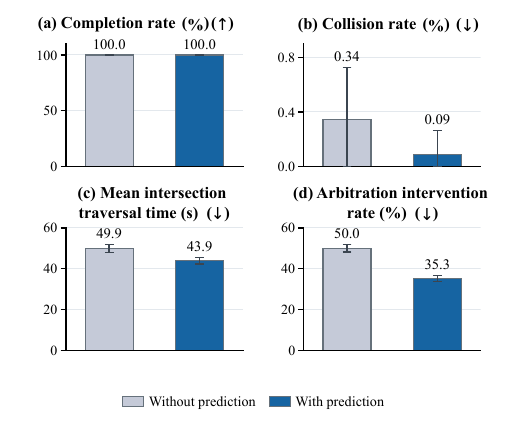}
\caption{HV-intent prediction ablation under the
hesitant-HV stress condition. Error bars denote 95\% CIs.}
\label{fig:hv_intent_ablation}
\end{figure}

The HV-intent ablation removes the inferred six-dimensional
features from HV graph nodes while retaining observed vehicle
states, CAV-intent exchange, and kinematic conflict checks.
As shown in Fig.~\ref{fig:hv_intent_ablation}, with 65\%
hesitant HVs, completion remains at 100\%, but the collision
rate rises from 0.09\% to 0.34\%, mean intersection traversal
time from 43.90 to 49.88~s. Arbitration intervention also increases from
35.27\% to 49.96\%.

HVs with similar current positions and speeds may respond
differently as they approach the conflict area. Features
inferred from their motion histories help the actor distinguish
these behaviors when generating raw intentions. Removing
these features leads to more arbitration corrections and
longer delays, although kinematic conflict screening remains
in place.

\begin{table}[!t]
\caption{Held-Out Performance of the HV-Intent Predictor}
\label{tab:hv_predictor_metrics}
\centering
\scriptsize
\setlength{\tabcolsep}{2pt}
\begin{tabularx}{\columnwidth}
{@{}>{\raggedright\arraybackslash}X
>{\raggedleft\arraybackslash}p{0.38\columnwidth}@{}}
\toprule
\textbf{Metric} & \textbf{Value}\\
\midrule
\multicolumn{2}{@{}l}{\emph{Behavior and movement/aggressive outputs}
($n=1{,}954$)}\\
Overall accuracy (\%) & 89.87\\
Balanced accuracy (\%) & 79.26\\
Aggressive-class recall (\%) & 66.47\\
Non-aggressive-class recall (\%) & 92.05\\
Movement-intent accuracy (\%) & 98.62\\
Movement-intent Brier score & 0.0110\\
Aggressive-intent accuracy (\%) & 89.92\\
Aggressive-intent Brier score & 0.0880\\
\midrule
\multicolumn{2}{@{}l}{\emph{Conflict-entry output}
($n=820$; 162 positive windows)}\\
Conflict-entry balanced accuracy (\%) & 81.30 [78.66, 83.16]\\
Conflict-entry recall (\%) & 88.89 [85.81, 92.35]\\
Conflict-entry AUROC (\%) & 90.44 [88.59, 92.06]\\
Conflict-entry AUPRC (\%) & 66.33 [58.76, 76.38]\\
Conflict-entry Brier score & 0.1018\\
\bottomrule
\end{tabularx}
\end{table}

The HV behavior-type classifier and the movement- and
aggressive-intent outputs are evaluated on 1,954
receiver-visible HV histories from 30 episodes and three
seeds. Balanced accuracy is the mean recall across classes,
whereas the Brier score measures the mean squared difference
between predicted probabilities and binary outcomes; lower
Brier scores are better. As reported in
Table~\ref{tab:hv_predictor_metrics}, the classifier achieves
89.87\% overall accuracy and 79.26\% balanced accuracy.
Aggressive-class recall is 66.47\%, below the 92.05\% recall
of the non-aggressive class. The movement- and
aggressive-intent outputs achieve accuracies of 98.62\%
and 89.92\%, with Brier scores of 0.0110 and 0.0880,
respectively.

The conflict-entry-intent output is evaluated on a separate
held-out set of 820 HV windows, including 162 entry events.
It achieves 81.30\% balanced accuracy, 88.89\% recall,
90.44\% AUROC, 66.33\% AUPRC, and a Brier score of 0.1018.
The recall indicates that most entry events within the
prediction horizon are identified; AUROC and AUPRC summarize
discrimination across decision thresholds. These direct
metrics describe predictive performance, while the preceding
ablation examines the contribution of the complete HV-intent
module to CAV decisions.

\FloatBarrier

\subsection{State-Channel Impairment Response}
\label{subsec:state_robustness}

The third evaluation objective first examines the trained
SIGN-MAPPO policy's robustness to state-channel impairment.
Vehicle-state delay is set to 30, 60, 90, 120, or 180~ms,
and packet loss ranges from 10\% to 50\%; CAV-intent
communication and all other settings remain unchanged.
Each impaired episode is paired with a zero-impairment
episode using the same environment and traffic seed. When a
state update is delayed or lost, the receiver uses its latest
valid record. Delays shorter than the 100-ms decision
interval are stochastically rounded to preserve their
expected values.

\begin{figure}[!t]
\centering
\includegraphics[width=\columnwidth]{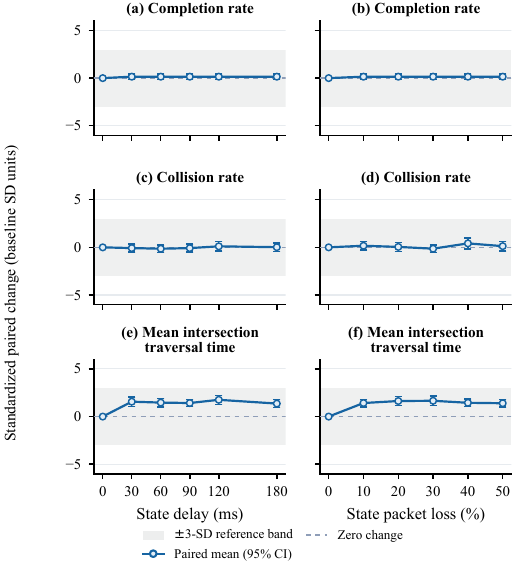}
\caption{State-channel robustness under delay (left column) and
packet loss (right column). Points and error bars denote standardized
paired mean changes and 95\% CIs relative to the corresponding
zero-impairment reference.}
\label{fig:state_channel_robustness}
\end{figure}

Because completion rate, collision rate, and mean intersection
traversal time differ in units and scale, their paired mean
changes are normalized by the episode-level standard deviation
under zero impairment:
\begin{equation}
z_m(\ell)=
\frac{\overline{\Delta m}(\ell)}{\sigma_m^0},
\label{eq:robustness_standardization}
\end{equation}
where $m$ denotes the metric, $\ell$ the impairment level,
$\overline{\Delta m}(\ell)$ the paired mean change from zero
impairment, and $\sigma_m^0$ the corresponding reference
standard deviation. The gray $[-3,3]$ region in
Fig.~\ref{fig:state_channel_robustness} is a visual reference
spanning three baseline standard deviations in each direction,
not an equivalence or safety bound.

Figure~\ref{fig:state_channel_robustness} shows that state
delay and packet loss have similar overall effects. In both
experiments, standardized changes in completion and collision
rates remain close to zero, while mean intersection traversal
time increases by approximately one to two baseline standard
deviations. Delayed or lost updates leave receivers using
retained peer states, which may alter estimated arrival order
and conflict-time overlap and postpone crossing decisions.
The packet-loss response is nonmonotonic because the effect
of a stale record depends on consecutive losses and the
vehicle's motion during that interval.

Receiver-specific reconstruction allows each CAV to continue
building its graph from the latest valid state records without
requiring a synchronized global view. CAV-intent exchange
remains available for coordination, and the graph policy and
local arbitration are reevaluated every 100~ms as state
information changes. These mechanisms are consistent with
the observed result: over the evaluated impairment ranges,
the main performance change is longer traversal time, while
completion and safety remain stable.

\FloatBarrier

\subsection{CAV-Intent Impairment under Nonideal State Sharing}
\label{subsec:intent_under_nonideal_state}

The CAV-intent delay and packet-loss experiments experiments extend the third evaluation
objective to conditions in which state sharing is already
nonideal, with state delay fixed at 60~ms and packet loss at
3\%. CAV-intent delay is varied from 30 to 180~ms with its
packet loss fixed at 1\%; CAV-intent packet loss is varied
from 10\% to 50\% with its delay fixed at 30~ms. The trained
policy and other settings remain unchanged. Each impaired
episode is paired with a reference using the same state-channel
setting and evaluation seed, with only the investigated
CAV-intent impairment set to zero. Paired changes are
standardized using (\ref{eq:robustness_standardization}).

\begin{figure}[!t]
\centering
\includegraphics[width=\columnwidth]
{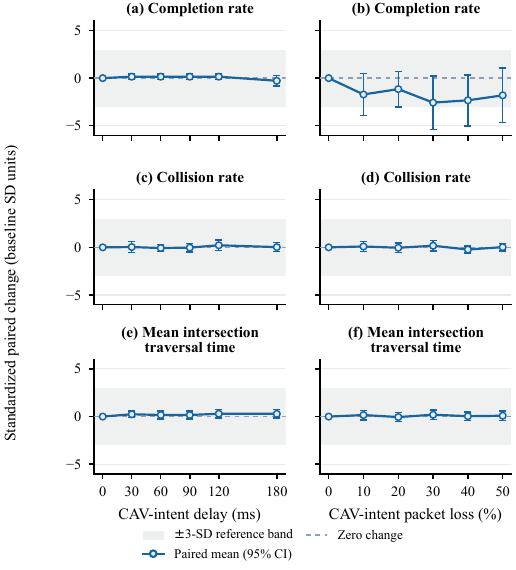}
\caption{Response to additional CAV-intent delay (left column)
and packet loss (right column) under fixed nonideal state
sharing. Points and error bars denote standardized paired mean
changes and 95\% CIs relative to the corresponding reference
with the same state-channel setting.}
\label{fig:intent_impairment_nonideal_state}
\end{figure}

As shown in the left column of
Fig.~\ref{fig:intent_impairment_nonideal_state}, increasing
CAV-intent delay produces only small changes in completion,
collision rate, and mean intersection traversal time. Their
standardized mean changes remain within approximately one
third of the reference standard deviations, without sustained
deterioration across the tested delays. A delayed message may
change the intentions available in one arbitration cycle, but
each CAV can use its retained records and reevaluates the
decision every 100~ms using its reconstructed traffic state.

The right column shows slight, nonmonotonic completion
fluctuations under additional CAV-intent packet loss, while
collision rate and traversal time remain close to the
reference. A lost request matters most when conflicting CAVs
seek access at the same time. The receiver retains its last
valid intention record, and arbitration still checks route
relations, state-derived priorities, and HV conflicts.
Compared with state-channel impairment, the additional
effects are smaller: state records affect graph features and
conflict estimates throughout decision making, whereas
received CAV intentions are used in coordinating crossing
requests. Across the evaluated ranges, completion, safety,
and passage efficiency therefore remain stable under the
combined communication impairments.

\FloatBarrier

\subsection{Cross-Scenario Evaluation}
\label{subsec:generalization}

As the final part of the third evaluation objective, the
cross-scenario evaluation aims to determine whether the fixed
SIGN-MAPPO policy maintains completion, safety, and passage
efficiency under changes in road scale, traffic demand, CAV
penetration, and HV behavior. The policy and communication
settings remain unchanged. G0 is the reference; G1--G8 vary
one factor at a time, while G9 combines high demand,
aggressive-HV dominance, and unpredictable HV maneuvers.
Table~\ref{tab:generalization} summarizes the results.

\begin{table}[!t]
\caption{Fixed-Policy Cross-Scenario Results}
\label{tab:generalization}
\centering
\scriptsize
\setlength{\tabcolsep}{0pt}
\begin{tabular*}{\columnwidth}
{@{\extracolsep{\fill}}lrrrr@{}}
\toprule
\textbf{Case} &
\textbf{Comp.} &
\textbf{Coll.} &
\textbf{Trav.} &
\textbf{Wait}\\
&
\textbf{(\%) $\uparrow$} &
\textbf{(\%) $\downarrow$} &
\textbf{(s) $\downarrow$} &
\textbf{(s) $\downarrow$}\\
\midrule
G0 Reference       & 100.00 & 0.000 & 21.92 & 4.06\\
G1 Two lanes       & 100.00 & 0.267 & 22.01 & 3.93\\
G2 Six lanes       & 100.00 & 0.087 & 18.05 & 4.43\\
G3 Low demand      & 100.00 & 0.185 & 15.19 & 1.75\\
G4 High demand     & 100.00 & 0.099 & 44.70 & 12.59\\
G5 25\% CAV        & 100.00 & 0.000 & 23.65 & 4.60\\
G6 75\% CAV        & 100.00 & 0.267 & 18.21 & 2.57\\
G7 Aggressive HV   & 100.00 & 0.356 & 19.69 & 3.55\\
G8 Hesitant HV     & 100.00 & 0.178 & 23.79 & 4.44\\
\midrule
G9 Compound stress & 100.00 & 8.420 & 43.07 & 13.24\\
\bottomrule
\end{tabular*}
\end{table}

Across G0--G8, completion remains at 100\% and collision
rates stay below 0.36\%, showing small changes in traffic
clearance and safety. Mean intersection traversal time varies
more noticeably. Lane-scale changes have relatively small
effects: G1 remains close to G0 (22.01 versus 21.92~s),
while G2 decreases to 18.05~s as arrivals are distributed
over more movement paths. Aggressive-HV-dominant G7 records
19.69~s, compared with 23.79~s in hesitant-HV-dominant G8.
Increasing CAV penetration from 25\% (G5) to 75\% (G6)
reduces traversal time from 23.65 to 18.21~s, consistent
with intention exchange being available to more vehicles.
Traffic demand has the largest effect: traversal time rises
from 15.19~s in G3 to 44.70~s in G4 as queues and
simultaneous route conflicts increase.

\begin{figure}[!t]
\centering
\includegraphics[width=\columnwidth]
{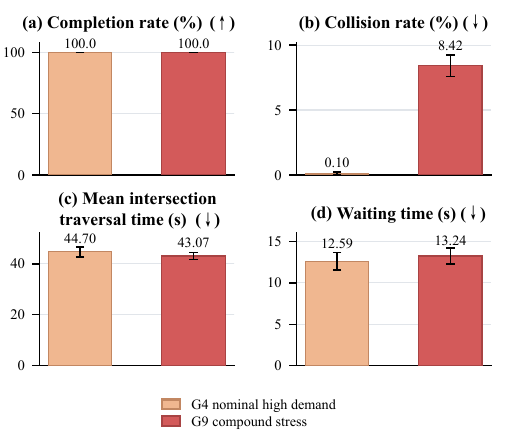}
\caption{Fixed-policy performance in the nominal high-demand
scenario G4 and the compound HV-behavior stress scenario G9.
Error bars denote 95\% CIs.}
\label{fig:generalization_stress}
\end{figure}

G9 combines high demand, 50\% CAV penetration, 70\%
aggressive HVs, and five unpredictable maneuvers: sudden
acceleration, emergency braking, forced lane changes,
following-vehicle rushes, and start--then--brake behavior.
One eligible HV is perturbed every 12--22~s. As shown in
Fig.~\ref{fig:generalization_stress}, G9 still achieves
100\% completion; its traversal and waiting times of 43.07
and 13.24~s remain close to G4's 44.70 and 12.59~s.
The collision rate rises from 0.10\% in G4 to 8.42\% in G9.
These repeated, abrupt maneuvers can change conflict-arrival
timing after the motion history used for a decision has been
observed, temporarily outdating the inferred intent and
conflict estimate. The fixed policy therefore retains traffic
clearance under this compound stress, although its safety
margin is reduced by the imposed short-notice HV behavior.

\FloatBarrier

\section{Conclusion}
\label{sec:conclusion}

This paper presented SIGN-MAPPO for cooperative CAV decision
making at mixed-traffic unsignalized intersections under
imperfect V2X communication. Each decision-eligible CAV
constructs a receiver-specific graph from its available
vehicle states and augments HV nodes with intent features
predicted from observed motion histories. A shared
graph-attention actor generates a raw \emph{Stop}/\emph{Go}
intention, which is exchanged through timestamped V2V
messages. Local arbitration then combines the available
CAV intentions with state-derived priorities and conflict
checks to determine the executable action. During training,
identity-indexed trajectories support shared-parameter MAPPO
updates under centralized training and decentralized
execution; the critic and parameter updates are omitted
during online execution.

The comparisons show a favorable balance of completion,
safety, and passage efficiency among the evaluated methods.
The ablations associate CAV-intent exchange with coordinated
release and HV-intent prediction with improved raw decisions
under uncertain HV motion. Communication tests show robust
performance over the evaluated delay and packet-loss ranges.
The fixed policy clears all generated vehicles across
G0--G9, with low collision rates across the G0--G8
single-factor scenarios. The evidence remains limited to
SUMO-based four-leg intersections and communication effects
modeled at the 0.1-s decision interval. Future work will
examine more diverse layouts, hardware-in-the-loop operation,
and measured asynchronous V2X traces.

\bibliographystyle{IEEEtran}
\bibliography{references}

\end{document}